%% file: neurips_2026.tex
\documentclass{article}

\usepackage{graphicx}
\usepackage{subcaption}
\usepackage{amsmath} 
\usepackage{wrapfig}
\usepackage{caption}
 \usepackage[dblblindworkshop, final]{neurips_2026}
\workshoptitle{Second Embodied Spatial Reasoning Workshop}
\usepackage[utf8]{inputenc}
\usepackage[T1]{fontenc}
\usepackage{hyperref}
\usepackage{url}
\usepackage{booktabs}
\usepackage{amsfonts}
\usepackage{nicefrac}
\usepackage{microtype}
\usepackage{xcolor}

\title{Aligning Policy Encoders with a Shape Teacher}

\author{%
  Madelon Bernardy \thanks{Corresponding author. Email: \texttt{madelon.bernardy@live.nl}} \\
  University of Amsterdam \\
  \And
  Mohammad Mahdi Derakhshani \\
  University of Amsterdam \\
  \And
  David M Knigge \\
  New Theory AI \\
  \And
  Riccardo Valperga \\
  University of Amsterdam \\
  \And
  Artem Moskalev \\
  New Theory AI \\
}

\begin{document}

\maketitle

\begin{abstract}
3D-aware representations can improve robotic manipulation by providing geometric information that is robust to changes in viewpoint, object position, and occlusion. However, using 3D inputs at deployment requires additional sensing and processing, making image-based policies more practical. We ask whether an image-based policy can acquire 3D shape information during training while retaining a purely image-based interface at deployment. We propose \textit{shape-teacher alignment}, a training-time mechanism that uses a pretrained 3D shape representation as an alignment target for an image-based policy encoder. Specifically, we use Point-MAE features from individual objects as targets for corresponding image patches, transferring geometric structure to the policy without changing its deployment inputs or architecture. On LIBERO, shape-teacher alignment produces image representations that are more structured around object shape and corresponding parts, while substantially accelerating policy convergence and improving success rates. For pretrained policies, convergence is accelerated by 4.7$\times$ and 5.6$\times$ on the Object and Spatial suites, respectively, without requiring the shape teacher or depth input at deployment. The project is available at: \url{https://miepjemadelon.github.io/aligningpolicieswithashapeteacher/}. 
\end{abstract}

\input{sections/introduction}
\input{sections/method}
\input{sections/results}
\input{sections/conclusion}
\bibliographystyle{plainnat}
\bibliography{references}

\newpage

\appendix
\section{Model Architecture}

Figure~\ref{fig:method_overview} shows a schematic overview of the method. The left side shows how the alignment target is determined, using back-projection to match point clusters to image patches. The right side depicts the action prediction pipeline.

\begin{figure}[h!]
    \centering
    \includegraphics[width=\linewidth]{images/method/method.pdf}
    \caption{Overview of the methodology. On the left, the pipeline to retrieve Point-MAE targets is visualized. On the right, the action-prediction pipeline is visualized.}
    \label{fig:method_overview}
\end{figure}

\section{Shape grouping Experimental Details}
For the shape grouping experiment, similar patches are grouped based on the object mask they land on. See Table~\ref {tab:clustering_objects} for the clustering groups. To get the final result, per encoded image, we select all patches that overlap by at least 50\% with object masks in Table~\ref{tab:clustering_objects}. For each of these patches in the object, the most similar patch in terms of cosine similarity is selected. Examples of this can be found in Figure~\ref{fig:patch_similarity} and in Figure~\ref{fig:dino}. We record whether this patch is inside the same object instance, inside an object within the same group (see Table~\ref{tab:clustering_objects}), or somewhere else in the image. We average these results over 1000 images to get the final results shown in  Table~\ref{tab:patch_similarity}.

\begin{table}[]
    \centering
    \caption{Similarity groups used for cluster analysis}
    \begin{tabular}{lr}
        \toprule
        Shape & Objects \\
        \midrule
         Drink Carton & Milk, Orange Juice  \\
         Box & Cream Cheese, Butter, Chocolate Pudding \\
         Tin & Tomato Sauce, Alphabet Soup \\
         Bottle & BBQ Sauce, Salad Dressing, Ketchup \\
         \bottomrule
    \end{tabular}
    \label{tab:clustering_objects}
\end{table}

\section{Iterations over Lambda}\label{sec:appendix_lambda}
\begin{figure}[!t]
    \centering
    \begin{minipage}{0.85\linewidth}
        \begin{subfigure}{0.49\linewidth}
            \centering
            \includegraphics[width=\linewidth]{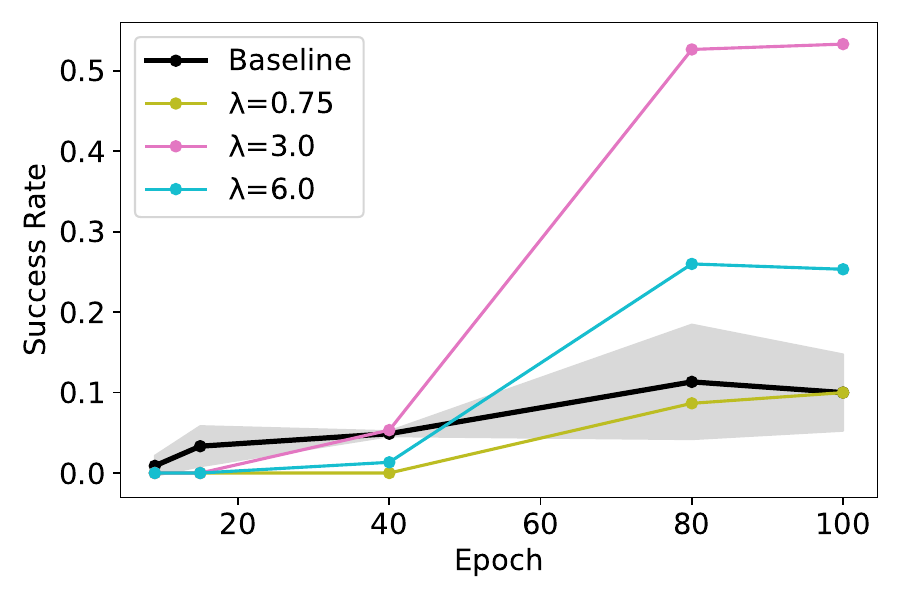}
            \caption{LIBERO-Object, from scratch.}
            \label{fig:lambda_no_pretrain}
        \end{subfigure}
        \hfill
        \begin{subfigure}{0.49\linewidth}
            \centering
            \includegraphics[width=\linewidth]{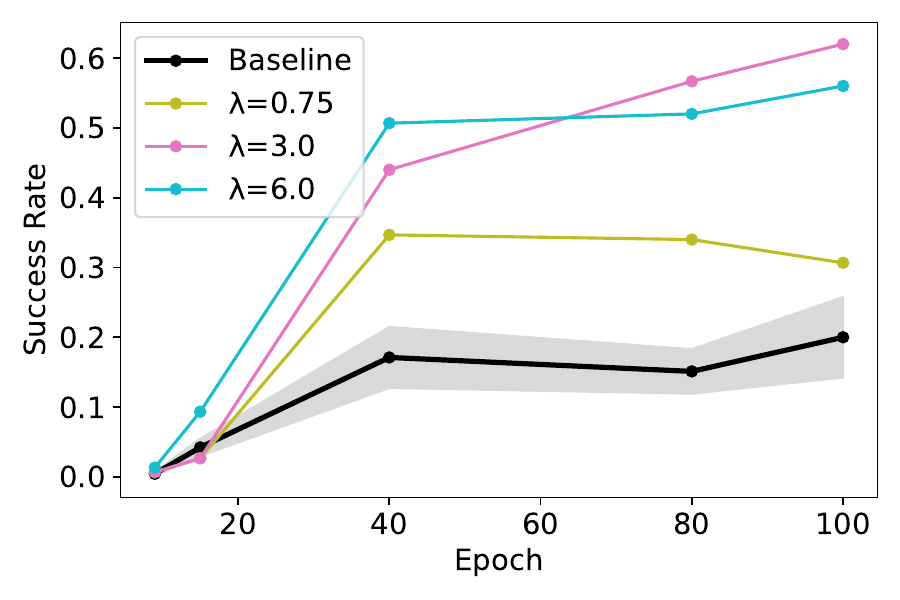}
            \caption{LIBERO-Spatial, from scratch.}
            \label{fig:lambda_no_pretrain_spatial}
        \end{subfigure}

        \vspace{0.4cm}

        \begin{subfigure}{0.49\linewidth}
            \centering
            \includegraphics[width=\linewidth]{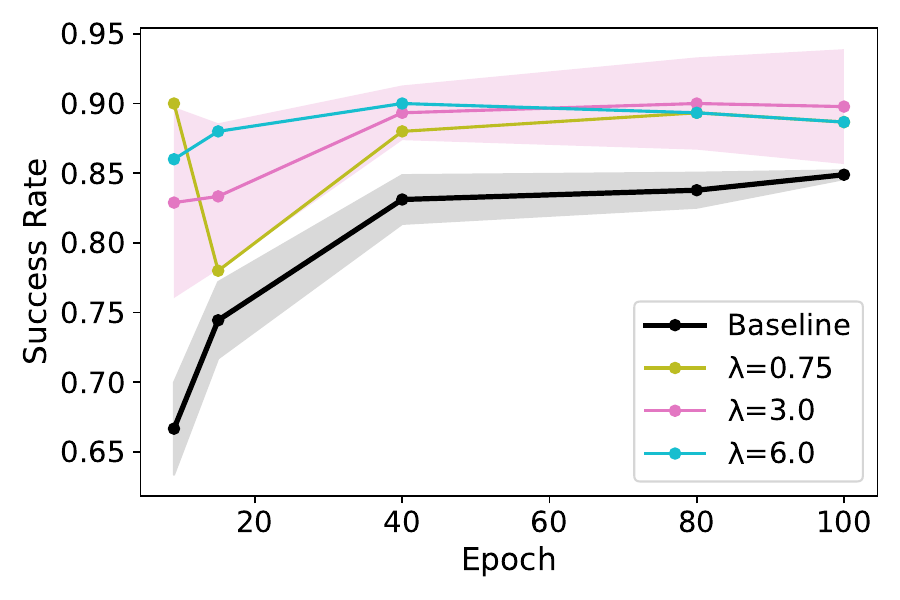}
            \caption{LIBERO-Object, pretrained.}
            \label{fig:target_object_uninit}
        \end{subfigure}
        \hfill
        \begin{subfigure}{0.49\linewidth}
            \centering
            \includegraphics[width=\linewidth]{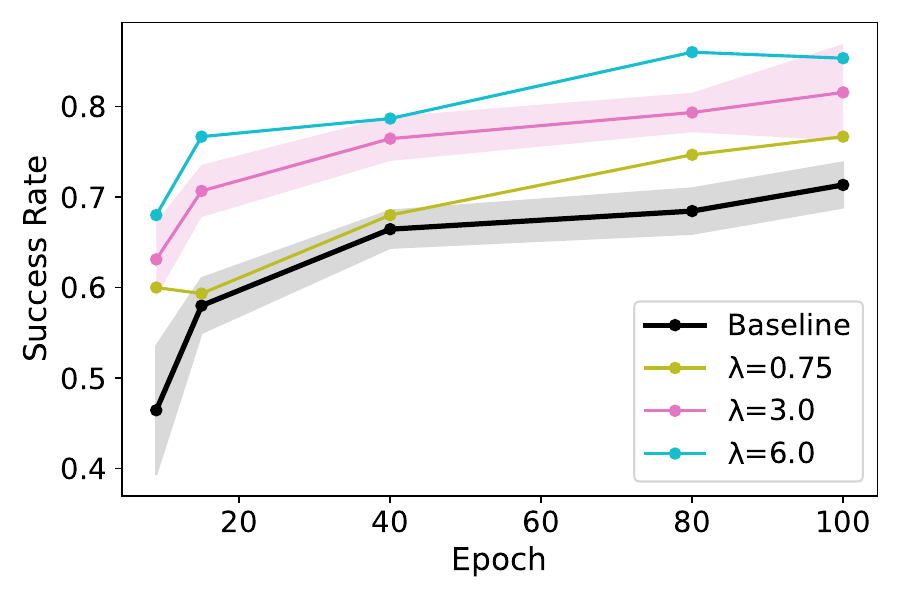}
            \caption{LIBERO-Spatial, pretrained.}
            \label{fig:target_spatial_uninit}
        \end{subfigure}
    \end{minipage}
    \caption{The convergence of each model on the LIBERO-Object and LIBERO-Spatial datasets for different alignment weights $\lambda$, for both pre-trained and from scratch configurations. Shaded regions show $\pm1$ standard deviation over three seeds.}
    \label{fig:lambda_convergence_pointmae}
\end{figure}

In addition to testing the effect of alignment for a fixed alignment weight, we also examined the effect of different alignment weights $\lambda$ on the model's convergence speed. The results for the from-scratch and pre-trained models can be found in Figure~\ref{fig:lambda_convergence_pointmae}. These results suggest that the alignment weight $\lambda$ has a sweet spot, and that exceeding it can hurt performance, especially when training from scratch, as shown on the LIBERO-Object set. However, the convergence results differ greatly per configuration. For instance, on the pre-trained models, a higher alignment weight does not seem to strongly affect convergence, nor does it clearly degrade convergence. Still, more configurations and more seeds are needed before we can draw a clear conclusion from these results.

\section{Additional Target Ablation Results}
\begin{table}[t!]
\caption{Percentage of query patches whose most similar patch falls on the same object, another object with similar shape, or an unrelated region, including DINOv2 \citep{oquab2024dinov2learningrobustvisual} results. For each query patch, the match is the patch with the highest cosine similarity among the evaluated images.}
\centering
\label{tab:patch_similarity_dino}
\begin{tabular}{lccc}
\toprule
 & Same object & Similar-shape object & Unrelated \\
\midrule
\multicolumn{4}{l}{\emph{From pre-trained}} \\
\hspace{1mm} Baseline & 10.2 & 8.2 & 81.6 \\
\hspace{1mm} Shape teacher aligned & 8.9 & 35.1 & 56.0 \\
\hspace{1mm} DINOv2 aligned & 31.7 & 5.7 & 62.6 \\
\midrule
\multicolumn{4}{l}{\emph{From scratch}} \\
\hspace{1mm} Baseline & 17.8 & 10.7 & 71.5 \\
\hspace{1mm} Shape teacher aligned & 1.3 & 36.4 & 62.2 \\
\hspace{1mm} DINOv2 aligned & 30.1 & 11.7 & 58.2 \\
\bottomrule
\end{tabular}
\end{table}

\begin{figure}[!t]
    \centering
    \begin{minipage}{0.85\linewidth}
        \begin{subfigure}{0.49\linewidth}
            \centering
            \includegraphics[width=\linewidth]{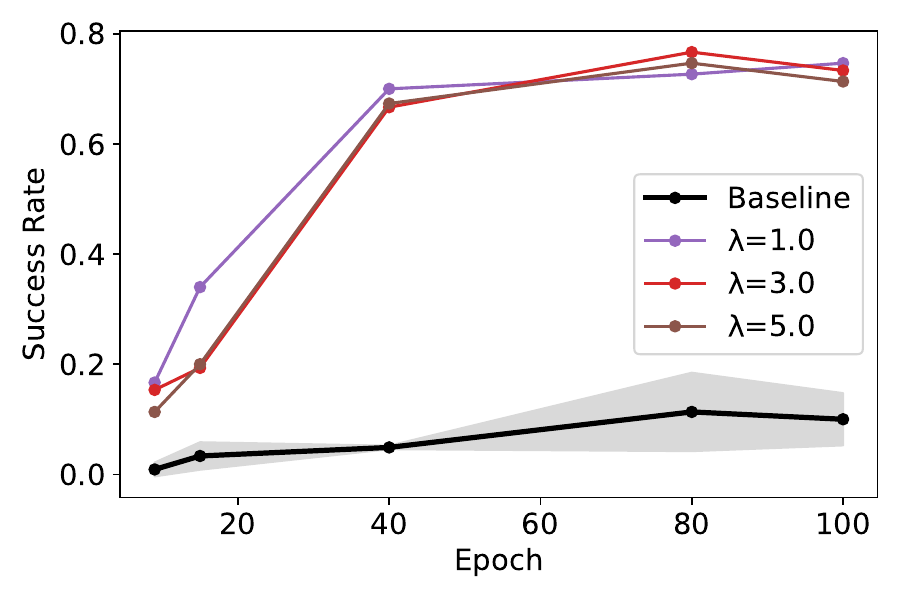}
            \caption{LIBERO-Object, from scratch.}
            \label{fig:lambda_no_pretrain}
        \end{subfigure}
        \hfill
        \begin{subfigure}{0.49\linewidth}
            \centering
            \includegraphics[width=\linewidth]{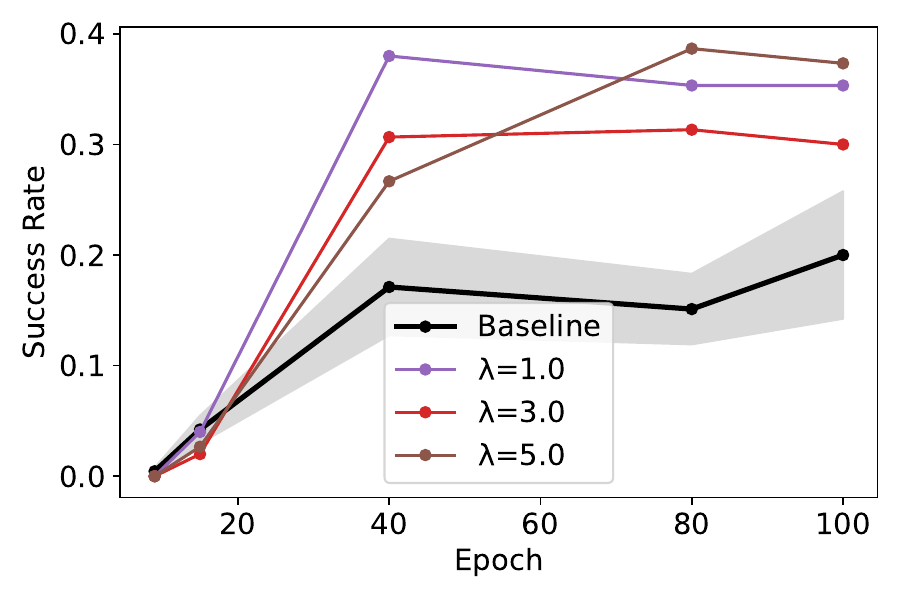}
            \caption{LIBERO-Spatial, from scratch.}
            \label{fig:lambda_no_pretrain_spatial}
        \end{subfigure}

        \vspace{0.4cm}

        \begin{subfigure}{0.49\linewidth}
            \centering
            \includegraphics[width=\linewidth]{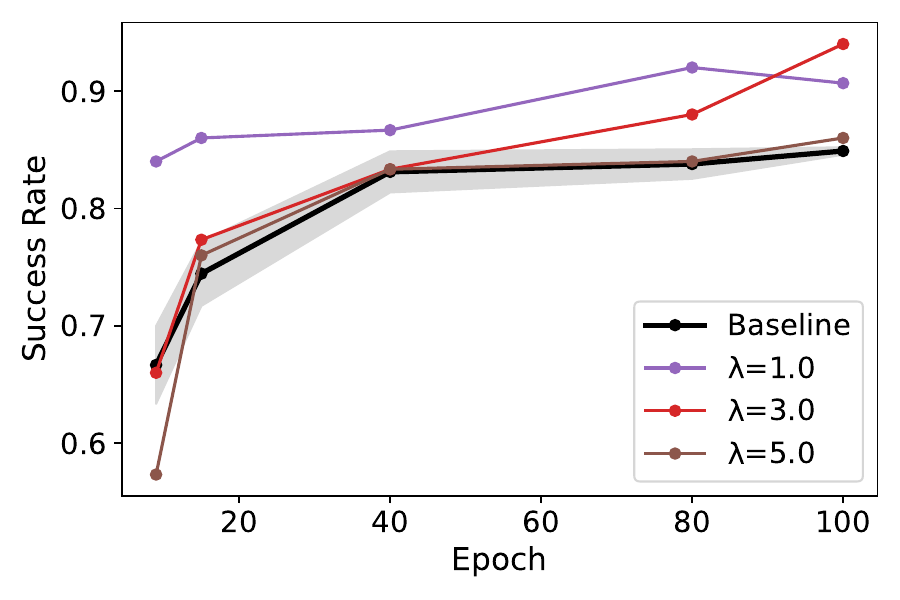}
            \caption{LIBERO-Object, pretrained.}
            \label{fig:target_object_uninit}
        \end{subfigure}
        \hfill
        \begin{subfigure}{0.49\linewidth}
            \centering
            \includegraphics[width=\linewidth]{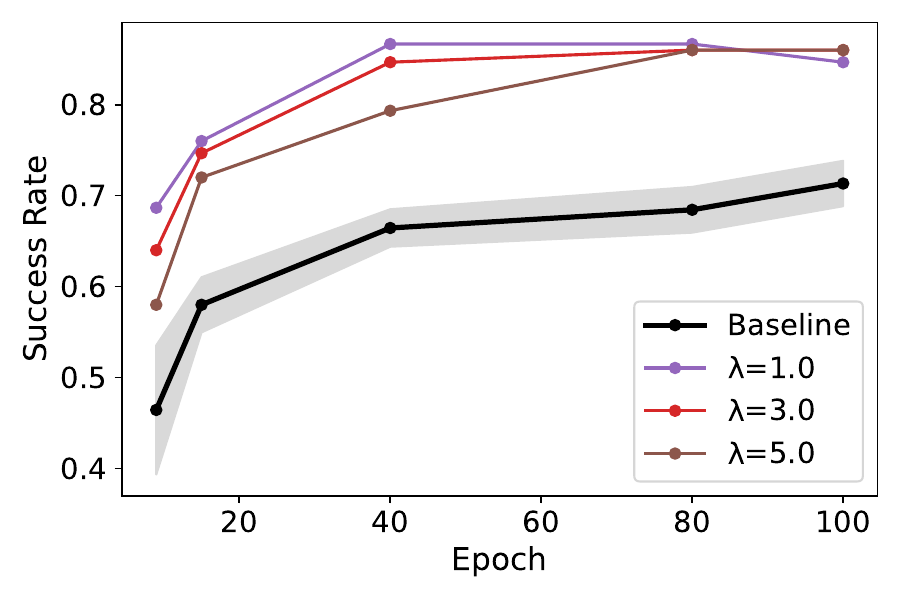}
            \caption{LIBERO-Spatial, pretrained.}
            \label{fig:target_spatial_uninit}
        \end{subfigure}
    \end{minipage}
    \caption{The convergence of each model on the LIBERO-Object and LIBERO-Spatial datasets for different alignment weights $\lambda$, for both pre-trained and from scratch configurations, for alignment with DINOv2. Shaded regions show $\pm1$ standard deviation over three seeds.}
    \label{fig:lambda_convergence_dino}
\end{figure}

We additionally compared PointMAE alignment against DINOv2 alignment for convergence speed, shown in Figure~\ref{fig:lambda_convergence_dino}, and for quantitative patch similarity, shown in Table~\ref{tab:patch_similarity_dino}. For convergence speed, our findings are similar to those discussed in Section~\ref{sec:appendix_lambda}: there appears to be an optimum value for the alignment weight, after which convergence speed drops or stays similar, depending on the dataset and on the pre-training. When considering the shape groupings of DINOv2, our findings again are similar to those in Section~\ref{sec:dino_ablation}. Where Point-MAE aligned models group patches with similar shapes, more so than patches within the same object, DINOv2 aligned models often select patches from the same object, rather than patches with the same shape. Overall, the fraction of patches whose most similar patch is on an unrelated item is similar for both models.

\section{Datasets}\label{chap:apx_datasets}
\begin{figure}[!ht]
    \centering
    \begin{subfigure}{\linewidth}
        \centering
        \includegraphics[width=0.85\linewidth]{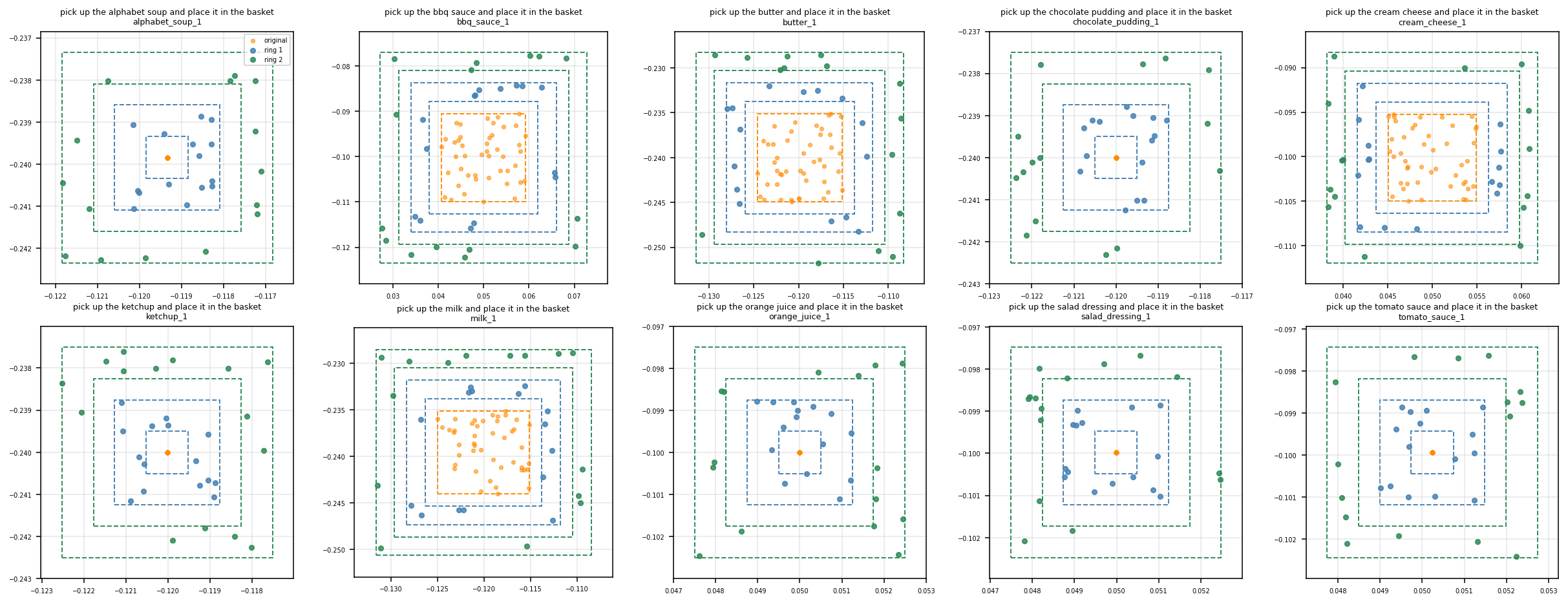}
        \caption{LIBERO-Object.}
        \label{fig:displacements_obj}
    \end{subfigure}

    \vspace{1em}

    \begin{subfigure}{\linewidth}
        \centering
        \includegraphics[width=0.85\linewidth]{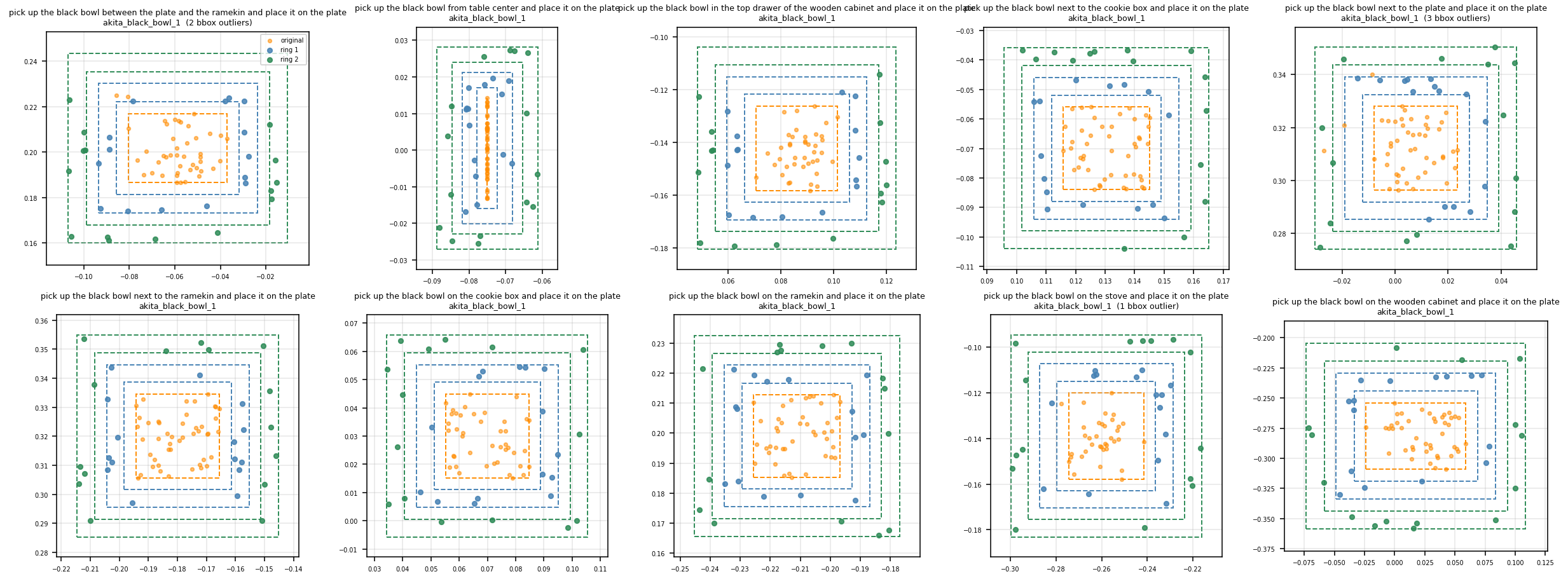}
        \caption{LIBERO-Spatial.}
        \label{fig:displacements_spatial}
    \end{subfigure}

    \caption{Displacements of the target object per task, for both suites. Orange marks the
    placements of the original dataset; blue and green mark the two out-of-distribution rings,
    ring~1 and ring~2, respectively.}
    \label{fig:displacements_rings}
\end{figure}

The out-of-distribution placements of the rings are shown in Figure \ref{fig:displacements_obj} and in Figure \ref{fig:displacements_spatial} for LIBERO-Object displacements and LIBERO-Spatial displacements, respectively. \\
To generate the rings, first, the object bounding box over the original dataset target displacements is computed. To generate LIBERO-Spatial displacements, outlier filtering was performed to generate the bounding box, as the spread is much larger than for LIBERO-Object. Not filtering out outliers caused performance on the outermost rings to drop to near zero. The Local Outlier Factor \citep{lof} was used to identify outliers.\\
From the bounding box, the rings are computed as fractions of the diameter of the bounding box. Ring 1 starts at $0.1\cdot \text{diameter}$ from the edge and ends at $0.25\cdot \text{diameter}$. Ring 2 starts at $0.35\cdot \text{diameter}$ from the edge and ends at $0.50\cdot \text{diameter}$. After the areas were created, 15 points were sampled from these areas using rejection sampling.

\section{Compute}
All experiments were run on the \textbf{\texttt{<redacted for anonymity>}} compute cluster. All experiments were run on an A100 GPU. Each training run took about 5 hours, and each checkpoint validation took about 15 minutes, for both standard and out-of-distribution validation. Experiments that only used the image encoder, such as patch similarities and t-SNE clustering, were negligible in compute, with each task taking less than 10 minutes. In total, training and evaluating the Point-MAE models, baseline models, and DINOv2 models used in the main paper, as well as the other experiments in the main paper, took around 150 hours on an A100 GPU. The hyperparameter search over $\lambda$ for both Point-MAE and DINOv2 took approximately 100 additional hours on an A100 GPU. However, the full research project required more resources, including additional configurations that did not make it into the paper and testing other properties of those models, such as different variants of alignment heads.

\end{document}

%% file: sections/introduction.tex
\section{Introduction}

Geometric information is important for robotic manipulation, with 3D-aware policies demonstrating improved data efficiency and robustness to variations such as viewpoint changes, object displacement, and occlusion \citep{3dvla,GeoVLA,pointvla,spatialvla,ze20243ddiffusionpolicygeneralizable,polarnet3d}. However, incorporating 3D observations into a robotic policy requires additional sensing and processing, including depth cameras, calibration, and depth estimation. These costs make purely image-based policies attractive in practice, but such policies can lack the geometric information needed for robust manipulation \citep{GeoVLA, gembench}. We therefore ask: can a 2D robotic policy acquire shape-aware representations during training without requiring 3D inputs at deployment?

Recent work on diffusion models has shown that pretrained representations can guide a target model through Representation Alignment (REPA). REPA \citep{yu2025representationalignmentgenerationtraining} aligns intermediate representations with features from a frozen pretrained encoder, improving convergence and generation quality, and has since been applied to robotic policies using pretrained vision encoders as alignment targets \citep{li2025spatialforcingimplicitspatial,agra}. However, these approaches do not specifically target 3D geometric structure, leaving open whether alignment with a representation that explicitly captures 3D shape can introduce geometric structure into an image-based policy. A shape teacher can encourage the policy encoder to represent corresponding geometric structure across objects, while the policy itself remains image-based at deployment.

We use a pretrained 3D shape teacher, Point-MAE \citep{avidan_masked_2022}, to align the image encoder of a 2D diffusion policy during training. Point-MAE features from individual objects are aligned with their corresponding image patches, encouraging the policy encoder to incorporate geometric structure while retaining the original image-based policy interface. We evaluate the resulting representations and policies on LIBERO, finding that shape-teacher alignment produces image representations that are more structured around object shape and corresponding object parts, while the resulting policies converge substantially faster and achieve higher success rates. For pretrained policies, convergence is accelerated by 4.7$\times$ and 5.6$\times$ on the Object and Spatial suites, respectively, with no need for the shape teacher or depth input at deployment. A comparison with a semantic teacher shows that shape and semantic alignment induce different representations, demonstrating that alignment can be used to impose distinct structural biases on the policy encoder.

To summarize, we make the following contributions:

\begin{itemize}
    \item We introduce shape-teacher alignment, a training-time mechanism for transferring 3D shape structure into an image-based robotic policy encoder without requiring 3D inputs at deployment.
    \item We instantiate shape-teacher alignment by matching per-object features from a pretrained 3D shape encoder to corresponding image patches, while keeping the deployed policy unchanged.
    \item On LIBERO, shape-teacher alignment produces more shape-structured representations and substantially faster policy convergence, with 4.7$\times$ and 5.6$\times$ speedups for pretrained policies on the Object and Spatial suites, respectively.    
\end{itemize}

%% file: sections/method.tex
\section{Method}

\paragraph{Baseline Diffusion Policy} \label{method:baseline}
We use the diffusion policy of \citet{chi_diffusion_2023} as our behavioral cloning baseline. The policy conditions on two consecutive observations, $t-1$ and $t$, and predicts a 10-step action trajectory, of which 8 steps are executed. We keep the action diffusion process unchanged, while modifying the observation encoder and transformer decoder. The observation encoder is a ViT/S-14 with the same architectural modifications as DINOv2: the CLS token is discarded, and the remaining patch tokens are used as observation features. The language token is produced by a frozen CLIP language model \citep{clip}, and the diffusion timestep is encoded using a sinusoidal embedding. The transformer decoder follows the cross-attention architecture of \citet{chi_diffusion_2023} and consists of a 12-layer decoder and a 2-layer encoder for the conditioning tokens. Inspired by \citet{ditblock}, the output projections of the cross-attention blocks are zero-initialized to improve stability.

\paragraph{Shape-Teacher Alignment} We align image-patch representations with features from a frozen pretrained 3D shape encoder. We use Point-MAE \citep{avidan_masked_2022} as the shape teacher, which encodes individual 3D objects into local shape features. For each object, we construct a point cloud from its mask and depth map, resample it to 1024 points, and partition the points into local groups of $M=32$ points. Point-MAE then produces a feature for each group.

The key challenge is to associate these 3D features with the corresponding image patches. We therefore back-project each 3D point to its corresponding pixel in the image and use these correspondences to determine how strongly each local 3D group contributes to each image patch (Figure \ref{fig:method_overview}). Let $B$ denote the set of object masks in an image, $P$ the set of image patches, and $A^j$ the set of point groups belonging to object $j\in B$. For a point group $a\in A^j$ and image patch $p\in P$, let $\pi(d,p)\in\{0,1\}$ indicate whether point $d\in a$ falls into patch $p$. The fraction of points in group $a$ that fall into patch $p$ is then
\begin{equation}
    \phi(a,p)=\frac{1}{M}\sum_{d\in a}\pi(d,p),
\end{equation}
which we use to weight the contribution of each 3D group to the patch. The total coverage of patch $p$ by object $j$ is
\begin{equation}
    \omega_p^j=\sum_{a\in A^j}\phi(a,p).
\end{equation}

Let $z_a^j$ denote the Point-MAE feature of group $a$ from object $j$. We use the coverage weights to aggregate the corresponding 3D features into a single alignment target for each image patch:
\begin{equation}
    t_p^j=
    \frac{1}{\omega_p^j}
    \sum_{a\in A^j}
    \phi(a,p) z_a^j.
\end{equation}
This target represents the 3D shape information associated with the portion of object $j$ visible in patch $p$.

We then align the corresponding image and 3D representations using a coverage-weighted cosine loss. Let $f_p$ denote the image encoder feature for patch $p$, and let $h$ be an alignment head with the same architecture as \citet{li2025spatialforcingimplicitspatial}. The alignment loss is
\begin{equation}
    \mathcal{L}_{align}(f)=
    \frac{
    \sum_{j\in B}\sum_{p\in P}
    \omega_p^j
    \left(1-\mathrm{cos\_sim}(h(f_p),t_p^j)\right)
    }{
    \sum_{j\in B}\sum_{p\in P}\omega_p^j
    }.
\end{equation}
Following REPA \citep{yu2025representationalignmentgenerationtraining}, we add the alignment loss to the behavioral cloning objective:
\begin{equation}
    \mathcal{L}=\mathcal{L}_{bc}+\lambda\mathcal{L}_{align},
\end{equation}
where $\lambda$ controls the strength of the alignment objective. The shape teacher is kept frozen and used only during training, so the deployed policy remains unchanged and incurs no additional inference cost.

%% file: sections/results.tex
\section{Experiments}

We evaluate whether shape-teacher alignment changes the learned image representations and improves policy learning, while preserving the image-only deployment interface. We consider two common regimes for robotic policy learning: training a task-specific policy from scratch on the target tasks, and fine-tuning a policy from a pretrained checkpoint. This allows us to evaluate shape-teacher alignment both when the policy learns its visual representation from scratch and when it starts from an existing pretrained representation.

\begin{wrapfigure}[20]{r}{0.38\columnwidth}
    \centering
    \vspace{-6mm}
    \includegraphics[width=\linewidth]{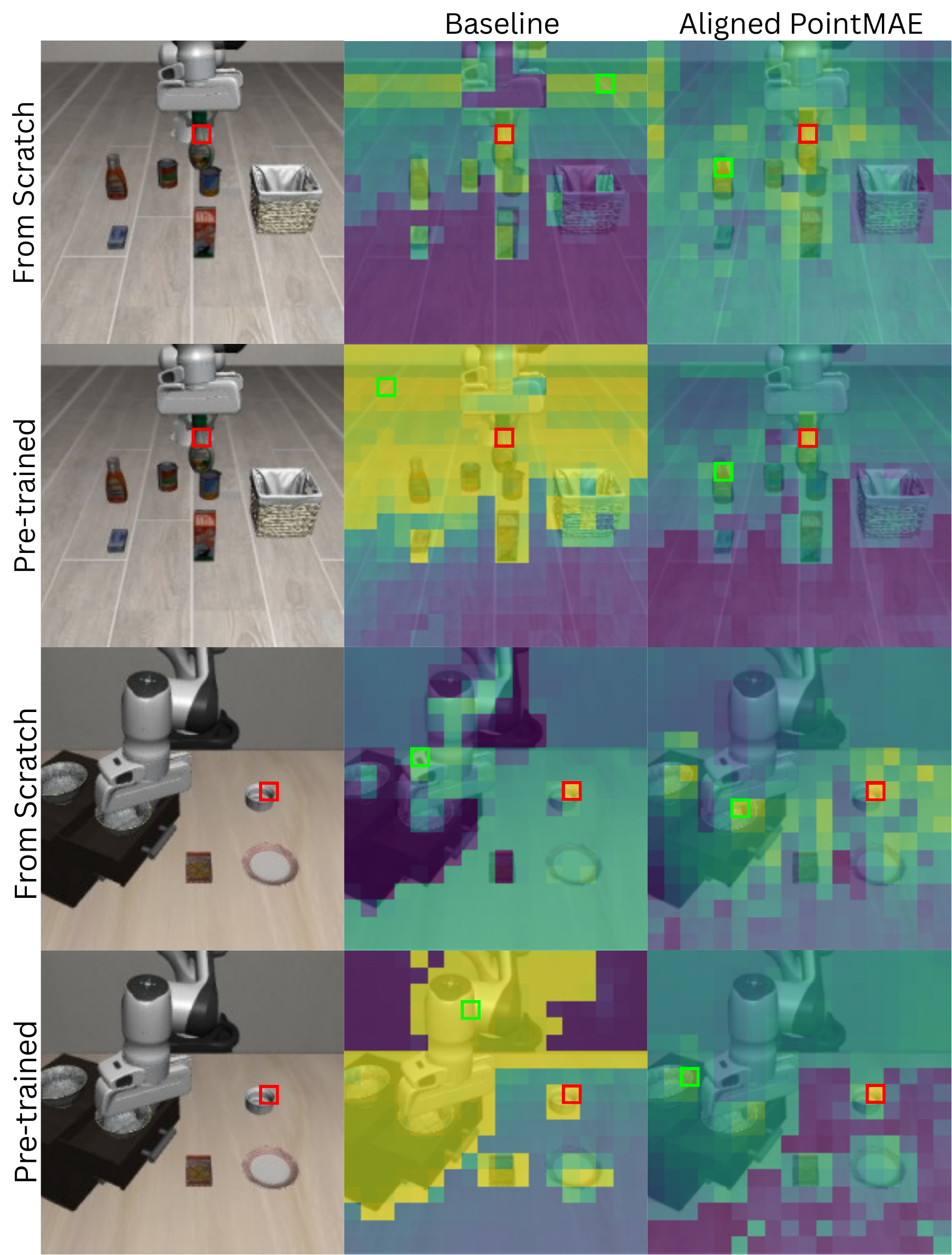}
    \caption{Patch similarity visualization. The red square marks the query patch and the green square its most similar patch.}
    \label{fig:patch_similarity}
\end{wrapfigure}

\paragraph{Dataset} We conduct all experiments on the LIBERO benchmark \citep{libero}, using the LIBERO-Object and LIBERO-Spatial suites. For experiments with pretrained policies, we use LIBERO-90 to pretrain the policy. To evaluate robustness to changes in object placement, we use Object-Displacement and Spatial-Displacement datasets constructed by placing targets on two rings around their original locations (Figure \ref{fig:displacements_rings}). We additionally evaluate zero-shot robustness to occlusion using LIBERO-Occ \citep{liberoocc}.

\paragraph{Evaluation} We evaluate whether shape-teacher alignment changes the structure of the learned image representations and improves downstream policy learning. We examine whether aligned features are organized more strongly by object identity, object category, shape, and corresponding object parts. Policy performance is measured by task success rate throughout training, allowing us to compare both final performance and convergence speed. For robustness experiments, we measure the change in success rate relative to the standard evaluation setting under target displacement and occlusion.

\paragraph{Baselines} We compare shape-teacher alignment against the corresponding no-alignment diffusion policy in both training regimes. In the \textit{from pretrained} setting, the policy is initialized from a checkpoint trained on LIBERO-90 and then fine-tuned on the target tasks. In the \textit{from scratch} setting, the same architecture is trained from random initialization on the target tasks. Unless otherwise stated, the aligned model uses Point-MAE as the shape teacher with alignment weight $\lambda=3$.
\begin{table}[t!]
\centering
\caption{Percentage of query patches whose most similar patch falls on the same object, another object with similar shape, or an unrelated region. For each query patch, the match is the patch with the highest cosine similarity among the evaluated images.}
\label{tab:patch_similarity}
\begin{tabular}{lccc}
\toprule
 & Same object & Similar-shape object & Unrelated \\
\midrule
\multicolumn{4}{l}{\emph{From pre-trained}} \\
\hspace{1mm} Baseline & 10.2 & 8.2 & 81.6 \\
\hspace{1mm} Shape teacher aligned & 8.9 & 35.1 & 56.0 \\
\midrule
\multicolumn{4}{l}{\emph{From scratch}} \\
\hspace{1mm} Baseline & 17.8 & 10.7 & 71.5 \\
\hspace{1mm} Shape teacher aligned & 1.3 & 36.4 & 62.2 \\
\bottomrule
\end{tabular}
\end{table}

\begin{figure}[b!]
    \centering

    \begin{subfigure}[t]{0.45\linewidth}
        \centering
        \includegraphics[
            width=\linewidth
        ]{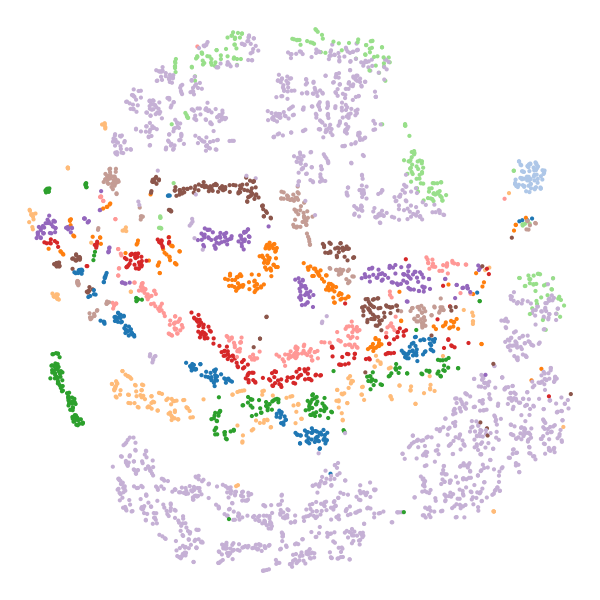}
        \caption{Baseline}
        \label{fig:tsne_baseline}
    \end{subfigure}
    \hfill
    \begin{subfigure}[t]{0.45\linewidth}
        \centering
        \includegraphics[
            width=\linewidth
        ]{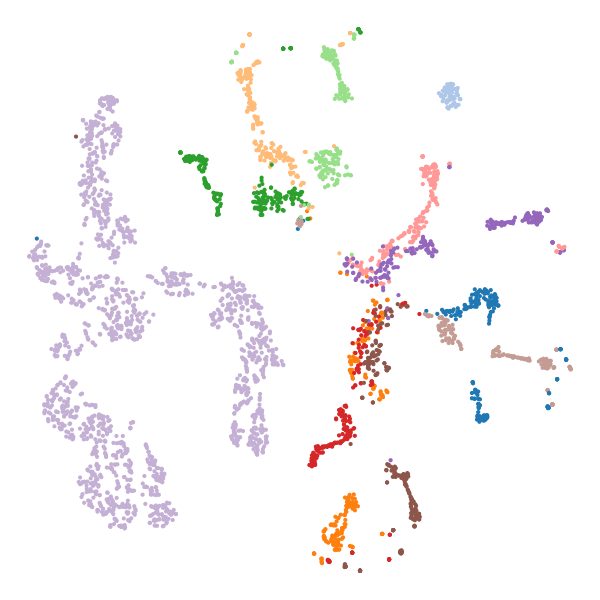}
        \caption{Shape-teacher aligned}
        \label{fig:tsne_aligned}
    \end{subfigure}

    \caption{Object-level image representations on LIBERO-Object for the from-scratch policy.\textit{ Colors correspond to object categories}. Shape-teacher alignment produces clearer object-level structure than the baseline.}
    \label{fig:tsne}
\end{figure}

\subsection{Shape- and Object-centric Representation}

We examine whether shape-teacher alignment transfers geometric structure into the image encoder, using patch-level similarity to assess correspondence across object parts and object-level clustering to assess object structure.

\paragraph{Shape Grouping} We first examine whether image patches representing similar object parts and shapes become more similar after alignment. For each selected patch, we identify the most similar patch among the image features and record whether it corresponds to the same object, another object with similar shape, or an unrelated region. Table~\ref{tab:patch_similarity} shows the resulting distribution on LIBERO-Object. For pretrained policies, alignment with the shape teacher increases the fraction of matches to another similar-shape object from 8.2\% to 35.1\%, while the fraction of matches falling into the ``unrelated'' category decreases from 81.6\% to 56.0\%. The same trend is observed for policies trained from scratch, where similar-shape matches increase from 10.7\% to 36.4\%. Figure~\ref{fig:patch_similarity} provides qualitative examples of these matches: aligned features can match corresponding parts across similarly shaped objects, such as the neck of a bottle or the side of a bowl, rather than only nearby patches of the same object. These results indicate that alignment introduces shape-related structure into the patch representations.

\paragraph{Object Grouping} We next examine whether alignment also organizes representations at the object level. We average the patch features corresponding to each object and visualize the resulting representations with t-SNE in Figure~\ref{fig:tsne}. Compared with the baseline, the aligned representations form clearer clusters of objects, indicating that alignment also produces stronger object-level structure. Together with the patch-level analysis above, this suggests that shape-teacher alignment organizes the image representation both across corresponding object parts and at the level of whole objects.

\begin{figure}[!t]
    \centering

    \begin{minipage}[t]{0.235\linewidth}
        \centering
        \includegraphics[width=\linewidth]{
            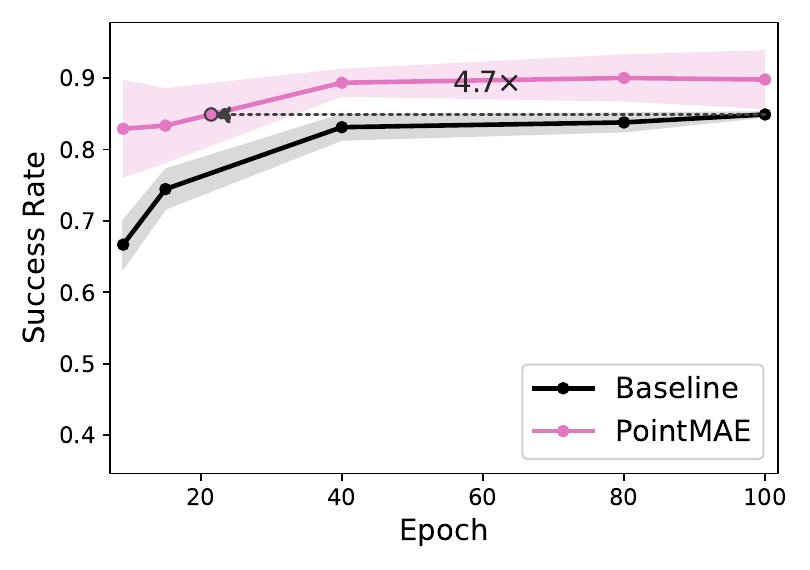
        }
        
        \vspace{1mm}
        \small LIBERO-Object, pretrained
    \end{minipage}
    \hfill
    \begin{minipage}[t]{0.235\linewidth}
        \centering
        \includegraphics[width=\linewidth]{
            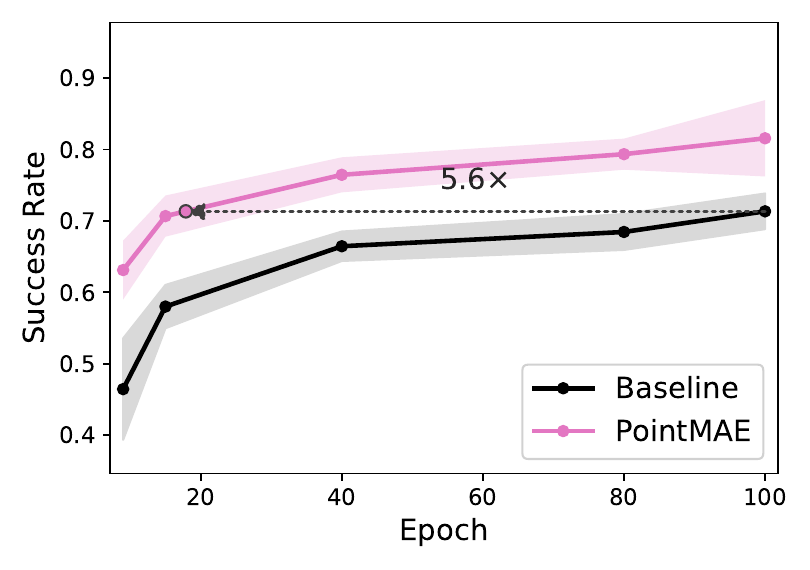
        }
        
        \vspace{1mm}
        \small LIBERO-Spatial, pretrained
    \end{minipage}
    \hfill
    \begin{minipage}[t]{0.235\linewidth}
        \centering
        \includegraphics[width=\linewidth]{
            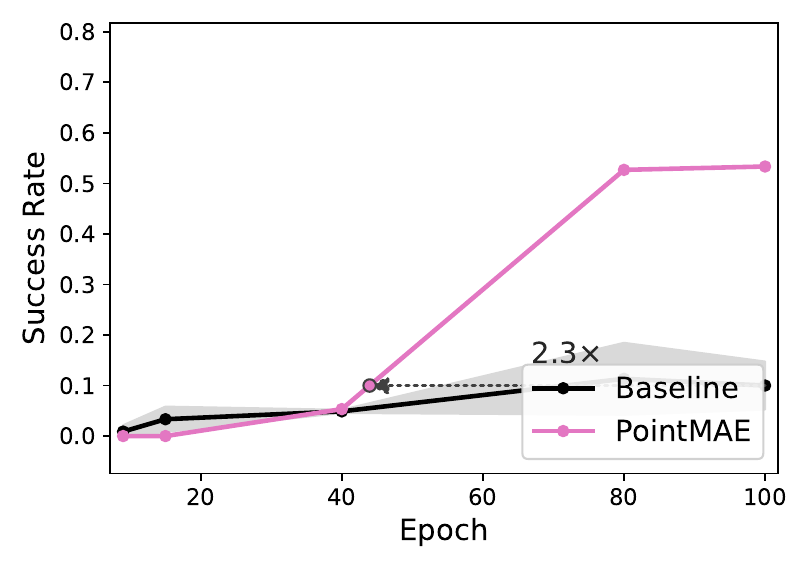
        }
        
        \vspace{1mm}
        \small LIBERO-Object, from scratch
    \end{minipage}
    \hfill
    \begin{minipage}[t]{0.235\linewidth}
        \centering
        \includegraphics[width=\linewidth]{
            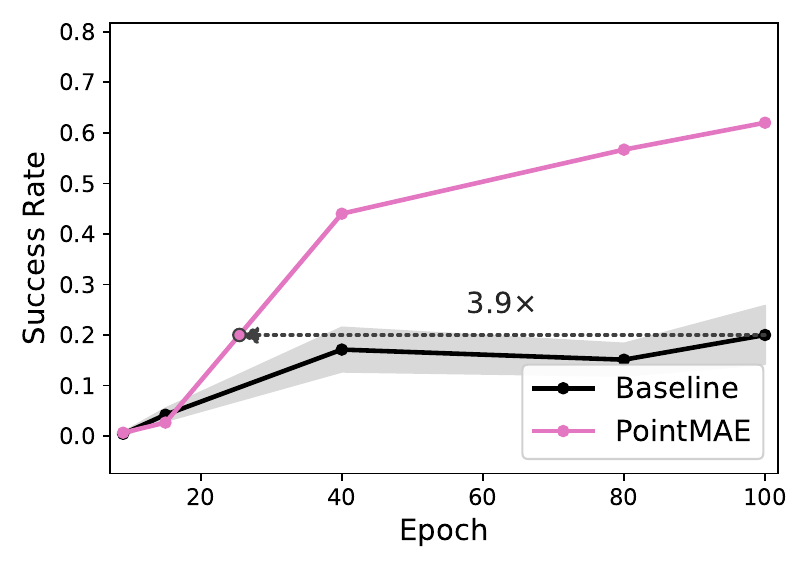
        }
        
        \vspace{1mm}
        \small LIBERO-Spatial, from scratch
    \end{minipage}

    \caption{Policy learning with and without shape-teacher alignment across LIBERO-Object and LIBERO-Spatial, for pretrained and from-scratch policies. Shape-teacher alignment accelerates convergence in both training regimes. Shaded regions show $\pm1$ standard deviation over three seeds.}
    \label{fig:convergence}
\end{figure}

\subsection{Improved Convergence}

We next examine whether the representational changes induced by shape-teacher alignment translate into faster policy learning. Figure~\ref{fig:convergence} compares the success rate throughout training for the baseline and Point-MAE-aligned policies in both training regimes. For pretrained policies, alignment reaches the same success rate as the baseline 4.7$\times$ faster on LIBERO-Object and 5.6$\times$ faster on LIBERO-Spatial. The effect is also present when training from scratch, with convergence accelerated by 2.3$\times$ and 3.9$\times$ on the Object and Spatial suites, respectively.

The consistent acceleration across both training regimes shows that the shape-teacher objective improves the optimization trajectory of the policy rather than only changing its final representation. The larger gains for pretrained policies suggest that shape-teacher alignment is particularly effective when adapting an existing visual representation to the target manipulation tasks.

\begin{figure*}[!t]
    \centering

    \begin{minipage}[t]{0.235\textwidth}
        \centering
        \includegraphics[
            width=\linewidth
        ]{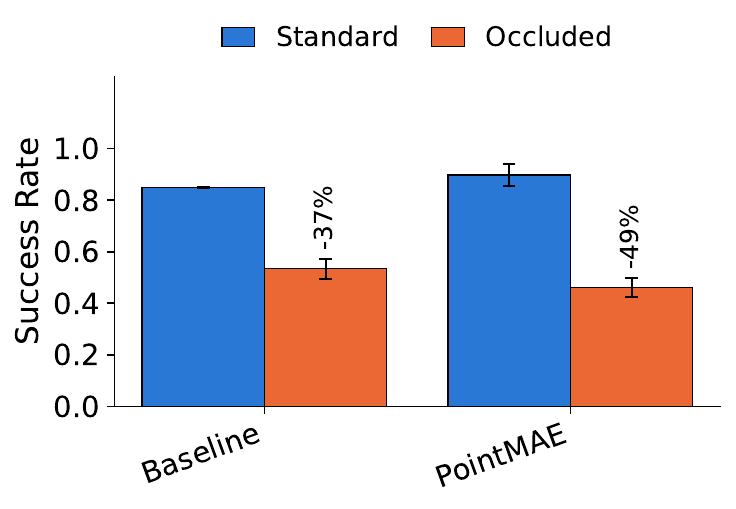}

        \vspace{1mm}
        \small LIBERO-Object, occluded
    \end{minipage}
    \hfill
    \begin{minipage}[t]{0.235\textwidth}
        \centering
        \includegraphics[
            width=\linewidth
        ]{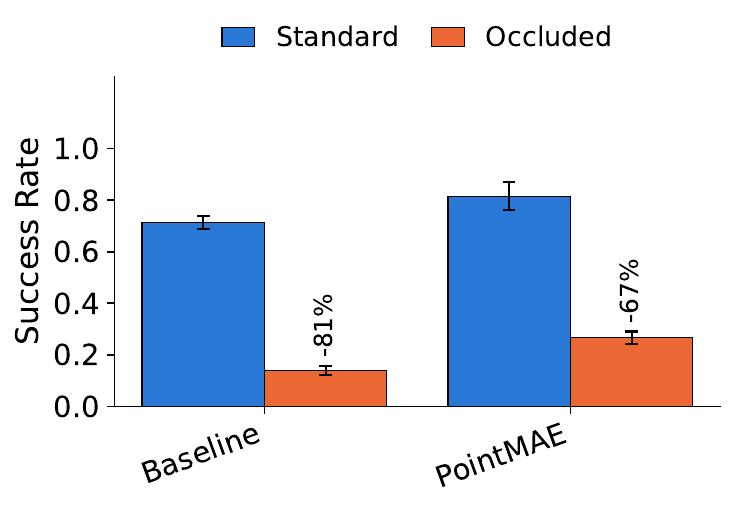}

        \vspace{1mm}
        \small LIBERO-Spatial, occluded
    \end{minipage}
    \hfill
    \begin{minipage}[t]{0.235\textwidth}
        \centering
        \includegraphics[
            width=\linewidth
        ]{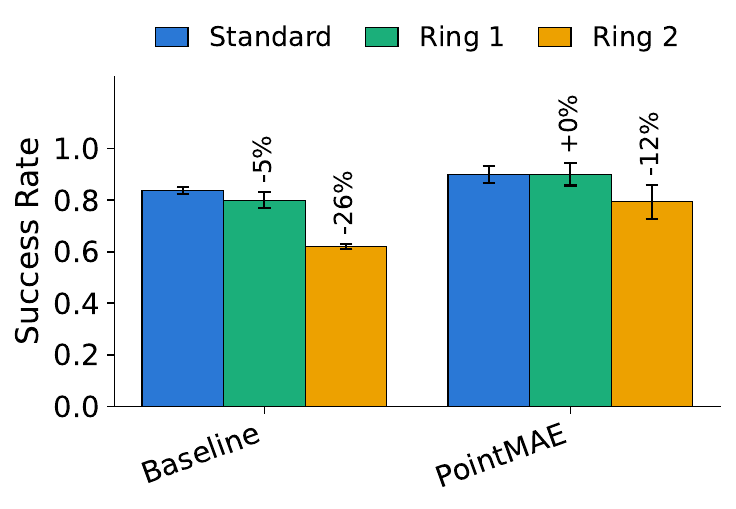}

        \vspace{1mm}
        \small LIBERO-Object, displaced
    \end{minipage}
    \hfill
    \begin{minipage}[t]{0.235\textwidth}
        \centering
        \includegraphics[
            width=\linewidth
        ]{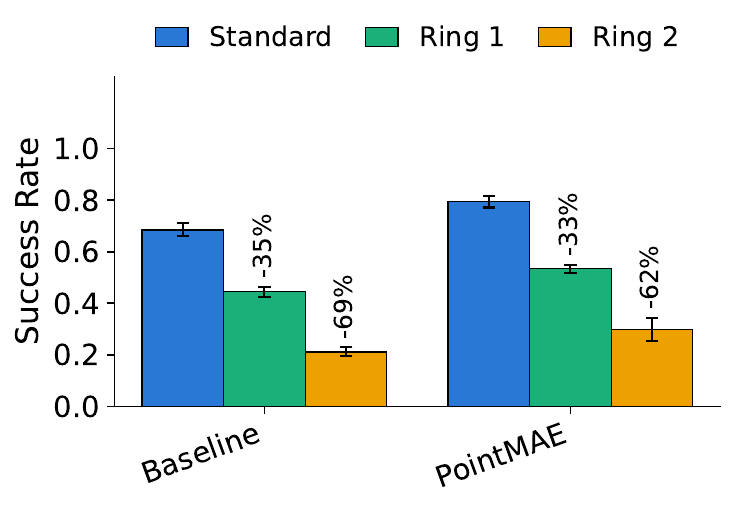}

        \vspace{1mm}
        \small LIBERO-Spatial, displaced
    \end{minipage}

    \caption{Zero-shot robustness to target occlusion and displacement for the pretrained policies. Annotations show the relative change in success rate from standard evaluation. Error bars show $\pm1$ standard deviation over three seeds.}
    \label{fig:robustness}
\end{figure*}

\subsection{Robustness}
We next examine whether the representational changes induced by shape-teacher alignment improve robustness to changes in the deployment environment. We evaluate performance under target displacement and object occlusion by measuring the change in success rate relative to the standard evaluation setting. Figure~\ref{fig:robustness} shows that the effect of alignment depends on both the type of distribution shift and the task suite. Under occlusion, the aligned policy has a similar performance drop to the baseline on LIBERO-Object but a smaller drop on LIBERO-Spatial. Under target displacement, alignment reduces the performance drop on LIBERO-Object, while the difference on LIBERO-Spatial is smaller. Thus, shape-teacher alignment does not provide a uniform improvement in robustness across all distribution shifts, although it performs similarly or better than the baseline across the tested conditions.

\subsection{Alignment Target Ablation}\label{sec:dino_ablation}

We finally examine whether the structure learned by the policy encoder depends on the choice of alignment target. We replace the Point-MAE shape teacher with DINOv2 \citep{oquab2024dinov2learningrobustvisual}, a pretrained semantic vision encoder, while keeping the alignment procedure and policy architecture unchanged. DINOv2-based runs use an alignment weight of $\lambda=1$. Table~\ref{tab:dino} compares the resulting policy performance and convergence speed. DINOv2 provides a strong alignment target and achieves higher success rates in several settings, while Point-MAE remains competitive on pretrained policies.

\begin{figure}[t]
    \centering
    \includegraphics[width=0.55\linewidth]{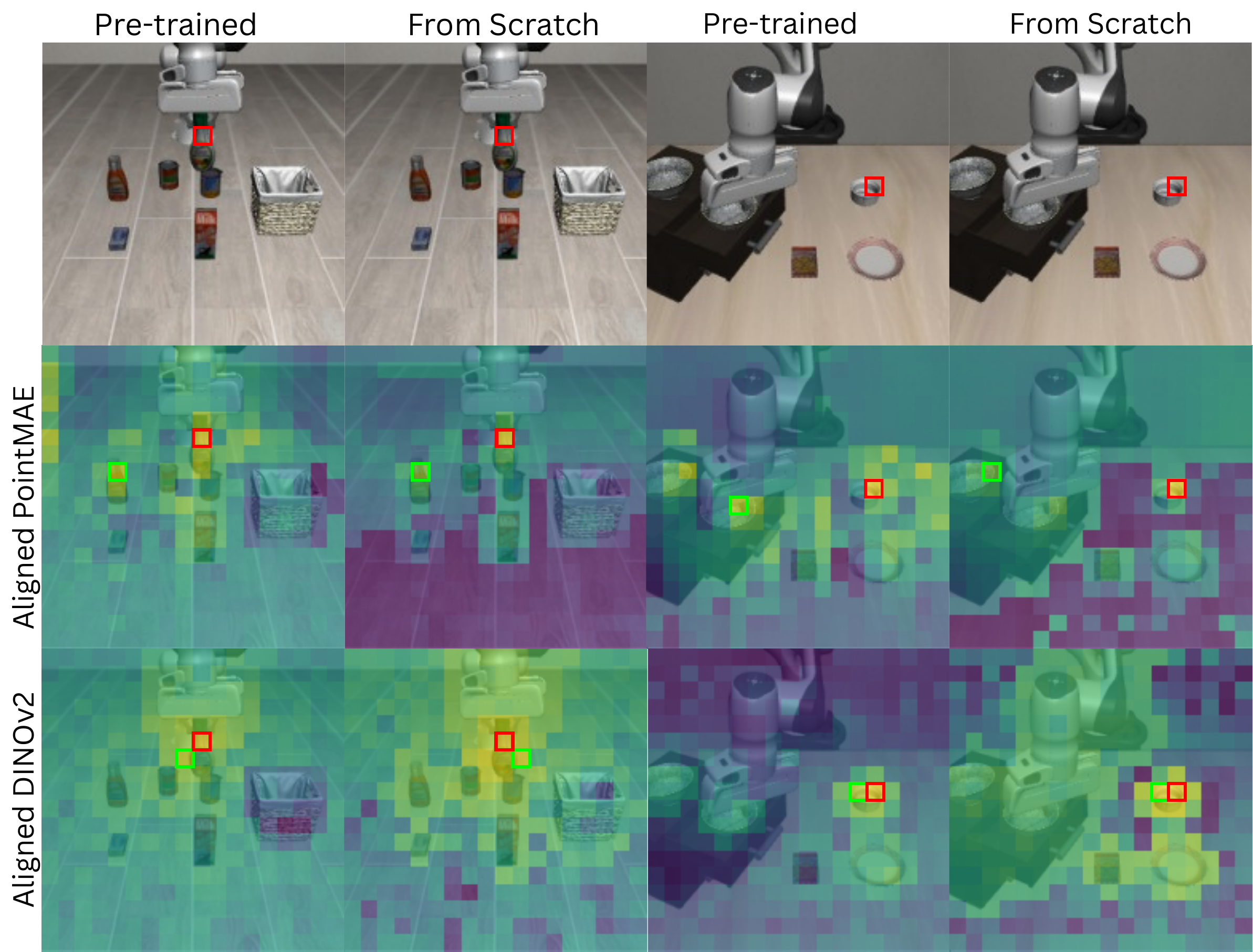}
    \caption{Patch-level similarity after alignment with DINOv2 and PointMAE. DINOv2 produces more object-oriented similarity patterns, whereas PointMAE produces greater similarity across corresponding parts of similarly shaped objects.}
    \label{fig:dino}
\end{figure}

More importantly, the two alignment targets produce qualitatively different representation structures. Figure~\ref{fig:dino} shows patch-level similarity for the two aligned encoders. DINOv2 produces more semantic-oriented similarity patterns, whereas Point-MAE produces greater similarity across similarly shaped objects and corresponding object parts. Thus, the effect of alignment is not determined only by the alignment procedure itself: the choice of teacher influences the structural bias learned by the policy encoder.

\begin{table}[t!]
    \centering
    \caption{Success rate at epoch 100 and convergence speed-up for alignment with PointMAE and DINOv2. Convergence speed-up is measured relative to the corresponding no-alignment baseline. Standard deviation over three seeds are reported for Baseline and PointMAE alignment.}
    \begin{tabular}{@{}llccc@{}}
    \toprule
    & & Baseline & PointMAE & DINOv2 \\
    \midrule
    Success rate (ep.\,100)
     & Object, from pretrained & $0.85 \pm 0.00$ & $0.90 \pm 0.04$ & $0.91$ \\
     & Object, from scratch & $0.10 \pm 0.05$ & $0.53$ & $0.75$ \\
     & Spatial, from pretrained & $0.71 \pm 0.02$ & $0.82 \pm 0.05$ & $0.85$ \\
     & Spatial, from scratch & $0.20 \pm 0.06$ & $0.62$ & $0.35$ \\
    \midrule
    Convergence speed-up
     & Object, from pretrained & -- & $4.7{\times}$ & $8.6{\times}$ \\
     & Object, from scratch & -- & $2.3{\times}$ & $\geq 11.1{\times}$ \\
     & Spatial, from pretrained & -- & $5.6{\times}$ & $8.9{\times}$ \\
     & Spatial, from scratch & -- & $3.9{\times}$ & $3.7{\times}$ \\
    \bottomrule
    \end{tabular}
    \label{tab:dino}
\end{table}

%% file: sections/conclusion.tex
\section{Conclusion}

We introduced shape-teacher alignment as a training-time mechanism for transferring 3D shape structure into an image-based robotic policy without requiring 3D inputs at deployment. On LIBERO, alignment produces image representations that are structured around both object shape and object identity, and substantially accelerates policy convergence, with the largest gains observed for policies initialized from pretrained checkpoints. Robustness to distribution shifts is more conditional, varying across perturbation types and task suites. Comparisons with a semantic teacher further show that the choice of alignment target determines the structure induced in the policy representation. These results suggest that representation alignment can be used not only to improve policy optimization, but also to deliberately shape the geometric structure learned by an image-based robotic policy while preserving its deployment interface.

\paragraph{Limitations and Future Work} Our evaluation is limited to a single policy architecture and the LIBERO benchmark, and uses Point-MAE as a single 3D shape teacher. The effect of shape-teacher alignment across other policy architectures, robotic manipulation benchmarks, and pretrained 3D representations therefore remains to be established. The robustness experiments also show that alignment does not uniformly improve performance under all distribution shifts. Future work will investigate whether stronger or more diverse 3D teachers, including representations that combine geometric and semantic information, can transfer complementary structure to the policy encoder, and whether alignment can improve robustness under more challenging forms of occlusion and distribution shift. Extending the evaluation to additional architectures and real-robot settings will further clarify the generality of training-time geometric alignment.